\pdfoutput=1

\documentclass[11pt]{article}

\usepackage{acl}

\usepackage{times}
\usepackage{latexsym}

\usepackage[T1]{fontenc}

\usepackage[utf8]{inputenc}

\usepackage{microtype}
\usepackage{graphicx}
\usepackage{subcaption} 
\usepackage{tabularx}
\usepackage{booktabs}
\usepackage{amsmath}
\usepackage{xspace}
\usepackage{xcolor}
\usepackage{enumitem}
\usepackage{stfloats}
\usepackage{xurl}

\newcommand{\hatecheck}{\textsc{HateCheck}\xspace}
\newcommand{\checklist}{\textsc{CheckList}\xspace}
\newcommand{\fscore}{F\textsubscript{1}\xspace}
\DeclareMathOperator*{\argmax}{argmax}
\DeclareMathOperator*{\argmin}{argmin}
\definecolor{myred}{HTML}{EB0C0C}
\definecolor{myblue}{HTML}{2F39CC}
\definecolor{mygreen}{HTML}{0EE22E}
\makeatletter
\newcommand\smaller{\@setfontsize\smaller{7.5}{8}}
\makeatother

\title{Checking \hatecheck: a cross-functional analysis of behaviour-aware learning for hate speech detection}

\author{Pedro Henrique Luz de Araujo \and
  Benjamin Roth \\
  University of Vienna \\
  \texttt{\{pedro.henrique.luz.de.araujo, benjamin.roth\}@univie.ac.at} \\ 
  }

\begin{document}
\maketitle
\begin{abstract}
Behavioural testing---verifying system capabilities by validating human-designed input-output pairs---is an alternative evaluation method of natural language processing systems proposed to address the shortcomings of the standard approach: computing metrics on held-out data. While behavioural tests capture human prior knowledge and insights, there has been little exploration on how to leverage them for model training and development. With this in mind, we explore behaviour-aware learning by examining several fine-tuning schemes using \hatecheck, a suite of functional tests for hate speech detection systems. To address potential pitfalls of training on data originally intended for evaluation, we train and evaluate models on different configurations of \hatecheck by holding out categories of test cases, which enables us to estimate performance on potentially overlooked system properties. The fine-tuning procedure led to improvements in the classification accuracy of held-out functionalities and identity groups, suggesting that models can potentially generalise to overlooked functionalities. However, performance on held-out functionality classes and i.i.d.\ hate speech detection data decreased, which indicates that generalisation occurs mostly across functionalities from the same class and that the procedure led to overfitting to the \hatecheck data distribution.
\end{abstract}

\section{Introduction}

The standard method for evaluating natural language processing (NLP) systems---computing metrics on held-out data---may be a good indicator of model correctness, but tends to overestimate performance in the wild~\cite{ribeiro_etal_checklist_ACL_2020}, does not indicate possible sources of models failure~\cite{wu_etal_errudite_acl_2019} and overlooks potential dataset biases~\cite{niven_kao_probeNLC_ACL_2019,mccoy_etal_right4wrongreasons_ACL_20219,zellers_etal_hellaSwag_ACL_2019}.

Behavioural testing of NLP models~\cite{rottger_etal_hateCheck_ACL_2021,ribeiro_etal_checklist_ACL_2020} has been proposed as an additional evaluation methodology, where system functionalities are validated by checking specific input-output behaviour of the system. This is done through challenge sets: expert-crafted input-output pairs that capture human prior knowledge and intuition about how an agent should perform the task~\cite{linzen_accelerateHLLG_ACL_2020} and enable systematic verification of system capabilities~\cite{belinkov_glass_analNLP_TACL_2019}.

For the purposes of this paper, we consider a behavioural test suite to be a collection of \textit{test cases}, input-output pairs that describe an expected behaviour. Each case assesses a specific \textit{functionality}, which are grouped into \textit{functionality classes}. For example, test cases in \hatecheck~\cite{rottger_etal_hateCheck_ACL_2021}, a test suite for hate speech detection, include (``[IDENTITY] belong in a zoo.'', hateful), (``No [IDENTITY] deserves to die.'', non-hateful) and (``I had this queer feeling we were being watched'', non-hateful). These cases assess the functionalities: \textit{implicit derogation of a protected group
or its members}, \textit{non-hate expressed using negated hateful statement} and \textit{non-hateful homonyms of slurs}\footnote{E.g., queer can be used as a slur for LGBT+ people, but also means strange, odd.}. These functionalities are grouped into the \textit{derogation}, \textit{slur usage} and \textit{negation} classes. A test suite may also contain \textit{aspects}, relevant properties of test cases that are orthogonal to the functionalities. An example of aspect in \hatecheck is the set of possible targeted identity groups.

While behavioural testing has been designed as a diagnostics tool, whether and how to leverage it for model training and development has seen little exploration, even though the human insights encoded in the test cases could potentially lead to more robust and trustworthy models. However, naively using behavioural testing for both training and evaluation is a risky affair---giving models access to the test cases could clue them into spurious correlations and lead to overestimation of model performance~\cite{linzen_accelerateHLLG_ACL_2020}. We view these risks as strong motivation to explore such settings, in order to gain insights into the vulnerability of behavioural tests to gaming and over-optimisation.

We explore three questions regarding behaviour-aware learning:

  \textbf{Q1}: Do models generalise across test cases from the same functionality? 
  This is a sanity check: test cases from the same functionality share similar patterns---sometimes generated by the same template---so we expect that behaviour-aware learning leads to better performance on test cases from functionalities seen during training.

  \textbf{Q2}: Do models generalise from covered functionalities to held-out ones? 
  By examining how behaviour-aware learning affects performance on held-out functionalities, we can estimate the robustness of the approach to potentially overlooked phenomena. 
  Equivalently, performance decrease is an indicator of overfitting to functionalities covered during training.

  \textbf{Q3}: Do models generalise from test cases to the target task?
  Improvements in the target task performance, as measured by independent and identically distributed (i.i.d.) data, would indicate that a model was able to extract the knowledge encoded in the behavioural tests.
  Conversely, a decrease in target task performance would signal overfitting to the behavioural test distribution.

In this paper, we explore behaviour-aware learning by fine-tuning pre-trained BERT~\cite{devlin_etal_bert_NACACL_2019} models on \hatecheck\footnote{Due to the nature of the task, this paper contains examples of abusive and hateful language. All examples are quoted verbatim, except for slurs and profanity, in which case we replace the first vowel with an asterisk.}. 
We experiment with several splitting methods and evaluate on different sets of held-out data: test cases for covered functionalities (\textbf{Q1}), test cases for held-out functionalities (\textbf{Q2}), and hate speech detection i.i.d.\ data (\textbf{Q3}).
In addition to \hatecheck's functionalities, we consider performance on held-out functionality classes and identity groups. By investigating our research questions, we address potential pitfalls and identify promising approaches for behaviour-aware learning\footnote{Our code is available on \url{https://github.com/peluz/checking-hatecheck-code}.}.

  \section{Related work}

  Traditional NLP benchmarks are created from text corpora assembled to reflect the naturally-occurring data distribution, which may fail to sufficiently capture important phenomena. Challenge sets were created as an additional evaluation framework, characterised by greater control over data that enables testing for specific linguistic phenomena~\cite{belinkov_glass_analNLP_TACL_2019}. 
\citet{ribeiro_etal_checklist_ACL_2020} proposed \checklist as a task-agnostic evaluation methodology with different test types that range from template-generated challenge sets to perturbation-based tests that enable checking behaviour on unlabelled texts. Inspired by \checklist,~\citet{rottger_etal_hateCheck_ACL_2021} created \hatecheck, a test suite for hate speech detection models composed of hand-crafted and template-generated test cases whose design was motivated by interviews with civil society stakeholders.

Using challenge data and behavioural tests to explicitly drive model development and training has largely gone unexplored. ~\citet{mccoy_etal_right4wrongreasons_ACL_20219} created HANS, a challenge set for natural language inference (NLI) designed to contradict classification heuristics that exploit spurious correlations in NLI datasets. They used the HANS templates to augment NLI training data, which helped prevent models from adopting such heuristics, though the improvement on held-out cases was inconsistent. 
\citet{liu_etal_InocFineTune_NACACL_2019} proposed inoculation by fine-tuning, where a model originally trained on a non-challenge dataset is fine-tuned on a few examples from a challenge set and then evaluated on both datasets. They do not assess generalisation from covered to held-out functionalities, as they use samples from the same functionality for training and testing.

To the best of our knowledge, we are the first to examine cross-functional behaviour-aware learning by fine-tuning models on different configurations of test suite and task data and evaluating performance across multiple generalisation axes.

\section{Cross-functional analysis of behaviour-aware learning}

We experiment with different training configurations by fine-tuning a pre-trained model on data from two distributions: the \textit{task} and the \textit{test suite}. The model is fine-tuned either on one of the distributions or on both sequentially, first on the task and then on the test suite. We compare the performance of the resulting models on both data distributions to assess the impact of behaviour-aware learning considering both task and challenge data.

Test suites have limited coverage: the included functionalities, functionality classes and aspects are only subsets of the phenomena of interest. For example, \hatecheck covers seven protected groups, which are particular samples of the full set of communities targeted by hate speech. Therefore, naive evaluation of models fine-tuned using test suite data can lead to overestimating their performance: models can overfit to the covered phenomena and pass the tests, but fail cases from uncovered phenomena (e.g., hate targeted at an uncovered identity group). Since we cannot directly evaluate performance on uncovered cases, we use performance on held-out sets of functionality, functionality classes and aspects as a proxy for generalisation across those three axes, as described in sections \ref{sec:testData} and \ref{sec:evaluation}.

\subsection{Task data} 
We use two hate speech detection datasets~\cite{davidson_etal_hateData_AAAIWeb_2017,founta_etal_hateData_AAAIWeb_2018} as source of task data.
Both are composed of tweets annotated by crowdsourced workers. 
The~\citet{davidson_etal_hateData_AAAIWeb_2017} dataset contains 24,783 tweets annotated as either hateful, offensive or neither, while the~\citet{founta_etal_hateData_AAAIWeb_2018} dataset contains 99,996 tweets annotated as hateful, abusive, spam or normal. 
We use the versions of the datasets made available\footnote{Available at \url{https://github.com/paul-rottger/hatecheck-experiments/tree/master/Data}.} by~\citet{rottger_etal_hateCheck_ACL_2021}, in which all labels other than hateful are collapsed into a single non-hateful label to match \hatecheck binary labels.
The data is imbalanced: hateful cases comprise 5.8\% and 5.0\% of the datasets, respectively.
We follow~\cite{rottger_etal_hateCheck_ACL_2021} and use a 80\%-10\%-10\% train-validation-test split for each of them. 

\subsection{Test suite data}
\label{sec:testData}
We use \hatecheck \cite{rottger_etal_hateCheck_ACL_2021} as the test suite. It contains 3,728 test cases that cover 29 functionalities grouped into 11 classes. \citet{rottger_etal_hateCheck_ACL_2021} created the set of functionalities based on interviews with 21 employees from NGOs that work with online hate. 18 of the functionalities deal with distinct expressions of hate, while the remaining 11 cover contrastive non-hate.
The test cases were either automatically generated using templates or created individually. We repeat the list of functionalities, classes and test case examples from \citet{rottger_etal_hateCheck_ACL_2021} in Appendix~\ref{sec:funcList}.

\citet{rottger_etal_hateCheck_ACL_2021} define hate speech as ``abuse that is targeted at a protected group or at its members for being a part of that group'', while protected groups are defined based on ``age, disability, gender
identity, familial status, pregnancy, race, national or ethnic origins, religion, sex or sexual orientation''. \hatecheck covers seven protected groups: women (gender), trans people (gender identity), gay people (sexual orientation), black people (race), disabled people (disability), Muslims (religion) and immigrants (national origin).  In addition to the gold label (hateful or non-hateful), each test is labelled with the targeted group.

When fine-tuning on test suite data, we use one of several splitting methods, as illustrated in Figure~\ref{fig:splitConfigurations}:

\begin{figure}[tb]
  \centering
  \includegraphics[width=\linewidth]{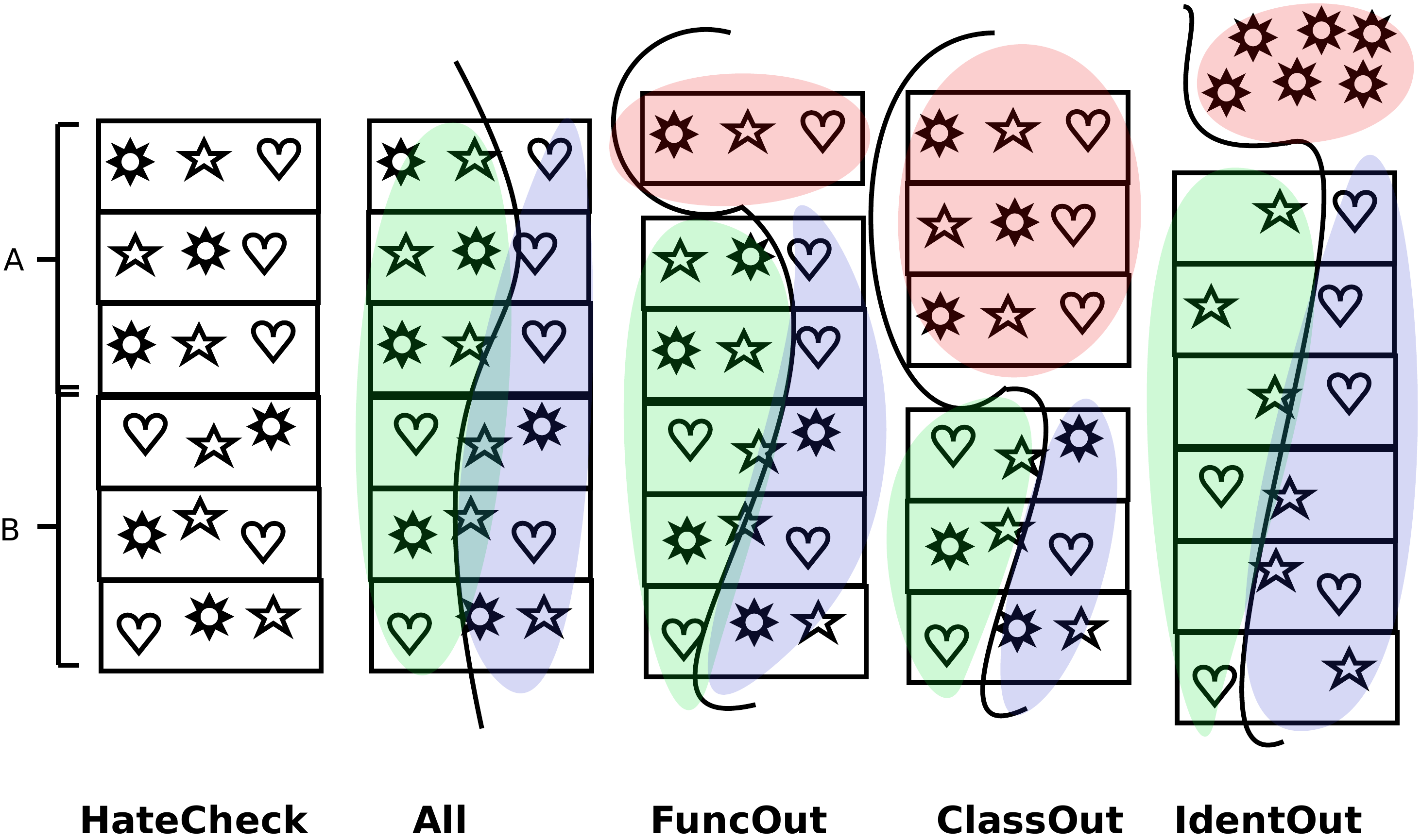}
  \caption{Illustration of our splitting techniques for \hatecheck.
  The first column shows a simplified version of \hatecheck with two functionality classes (A and B) that each contain test cases targeting three identity groups (denoted by suns, stars and hearts) grouped into three functionalities (denoted by the rectangles).
  In all splitting schemes, test cases are randomly split between {\color{teal} training} and {\color{myblue} evalu}{\color{myred} ation} sets, as indicated by the curved lines; 
  the difference lies in whether a set of test cases with specific properties not covered in training is {\color{myred} held-out} for evaluation.
  All split: no fixed set held out.
  FuncOut split: test cases from one functionality held out.
  ClassOut split: test cases from one functionality class held out.
  IdentOut split: test cases targeting a identity group held out. 
  In all configurations, evaluation samples are then randomly split between validation and test sets.}
  \label{fig:splitConfigurations}
\end{figure}%

\textbf{All} A random 50\%-25\%-25\% train-validation-test split. 

\textbf{FuncOut} 
We first hold out all test cases from a given functionality and randomly split the remaining cases into a 50\%-50\% train-evaluation split. 
We divide the union of held-out and evaluation split cases into a 50\%-50\% validation-test split.
The process is repeated for each functionality, resulting in 29 split configurations.

\textbf{IdentOut}
The same as FuncOut, but test cases relating to each identity group are held out, resulting in 7 split configurations.

\textbf{ClassOut} 
Similar to the previous two, but entire functionality classes are held out, resulting in 11 split configurations.

\subsection{Training configurations}
We consider the following training configurations:

\textbf{Task-only} Models are fine-tuned only on the task data. We denote the task-only configurations as Davidson and Founta, depending on which dataset was used for training.

\textbf{Test suite-only} Models are fine-tuned only on test suite data. We denote the test suite-only configurations by the name of the splitting method used.

\textbf{Task and test suite} Models are sequentially fine-tuned first on task data and then on test suite data. We denote these configurations as [Task data]-[Test suite split]. For example, in the Davidson-FuncOut configuration, models are first fine-tuned on the Davidson split and then on the FuncOut splits.

\subsection{Evaluation}
\label{sec:evaluation}

We evaluate the models that result from each training configuration on both task and test suite data. 
For task evaluation (\textbf{Q3}), due to the label imbalance, we report the macro \fscore score computed on Davidson or Founta test sets. 
For test suite evaluation, we follow \citet{rottger_etal_hateCheck_ACL_2021}, and use the accuracy as the classification metric. We measure generalisation to covered functionalities and identities (\textbf{Q1}) by computing the All test set performance. 

We aggregate performance on IdentOut test sets in the following way: for each of the seven IdentOut split configurations we fine-tune the model on the train split and use it to compute the {\color{myred}held-out} test predictions and the {\color{myblue} covered} test accuracy (Figure~\ref{fig:splitConfigurations}).
We compute the accuracy on the union of the seven held-out prediction sets as the held-out performance measure, and the average covered test accuracy as the covered performance measure\footnote{
  Covered and held-out aggregation methods are different because each of the seven held-out test sets targets a single identity group. Consequently computing the accuracy on each set and averaging them all would result in the average identity group accuracy instead of the overall test accuracy.}.
The same method is used to aggregate performance on FuncOut and ClassOut sets. 

The obtained held-out accuracies are measures of generalisation to held-out identity groups, functionalities and functionality classes (\textbf{Q2}). Additionally, FuncOut and ClassOut test sets are used to contrast generalisation to related (intra-class) and unrelated (extra-class) functionalities: in the former case, a model that has no access to \textbf{F14} (hate expressed using negated positive statement), will be trained on \textbf{F15} (non-hate expressed using negated hateful statement) cases; in the latter, there are no \textit{negation} samples in the train split.

\subsection{Experimental setting}
All models start from a pre-trained uncased BERT-base model\footnote{Model card available in \url{https://huggingface.co/bert-base-uncased}.}. When fine-tuning, we follow \citet{rottger_etal_hateCheck_ACL_2021} and use cross-entropy with class weights inversely proportional to class frequency as the loss function and AdamW~\cite{loschilov_hutter_adamw_ICLR_2019} as the optimiser. We also search for the best values for batch size, learning rate and number of epochs through grid search, selecting the configuration with the smallest validation loss.

\section{Results and discussion}

\begin{figure}[tb]
  \centering
  \includegraphics[width=\linewidth]{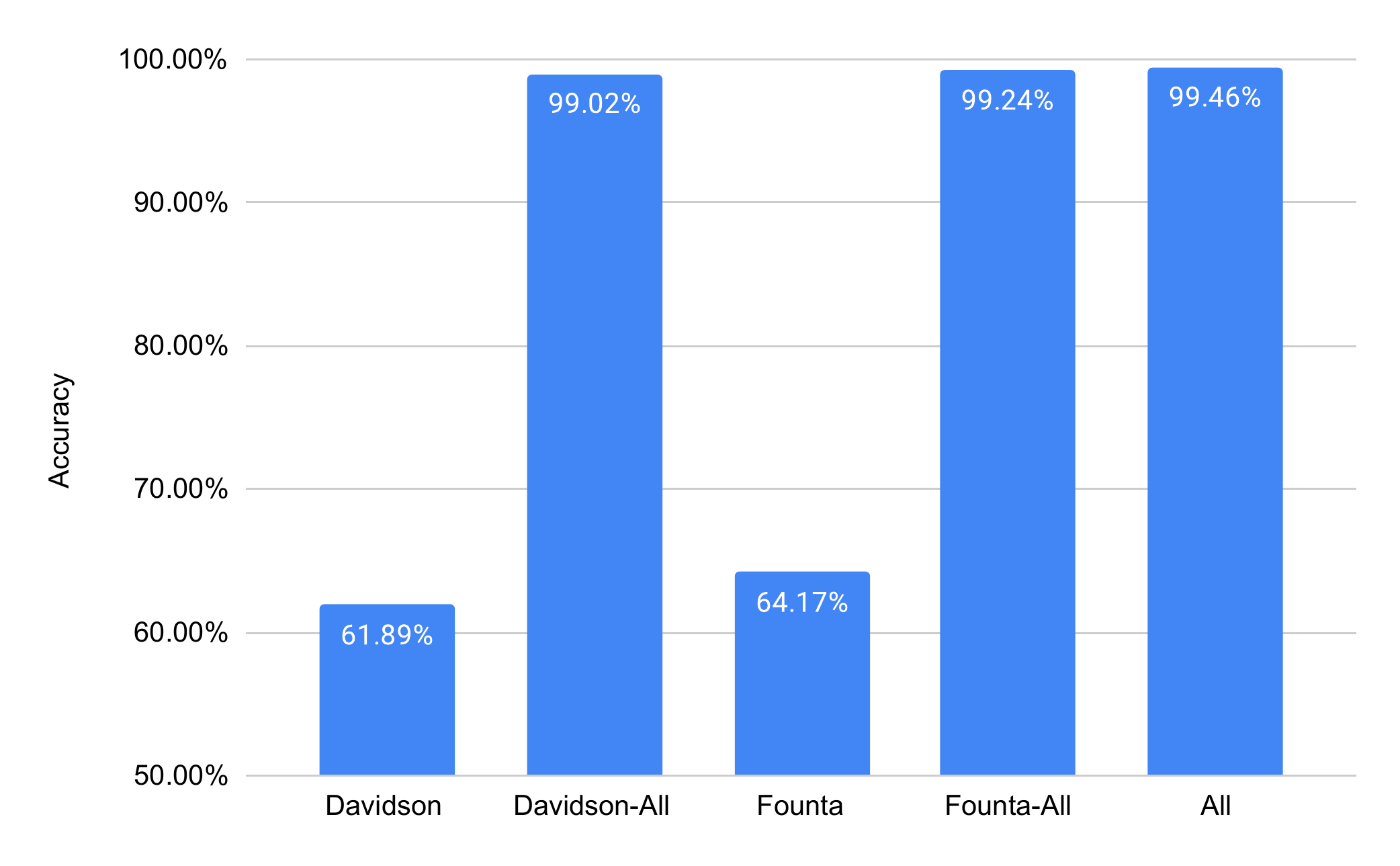}
  \caption{Performance on All split test set: models fine-tuned on \hatecheck outperform the ones trained only on task data.}
  \label{fig:AllSplitResults}
\end{figure}%

\begin{figure*}[tb]
  \begin{tabularx}{\linewidth}{XXX}
  \begin{subfigure}{\linewidth}\centering\includegraphics[width=\linewidth]{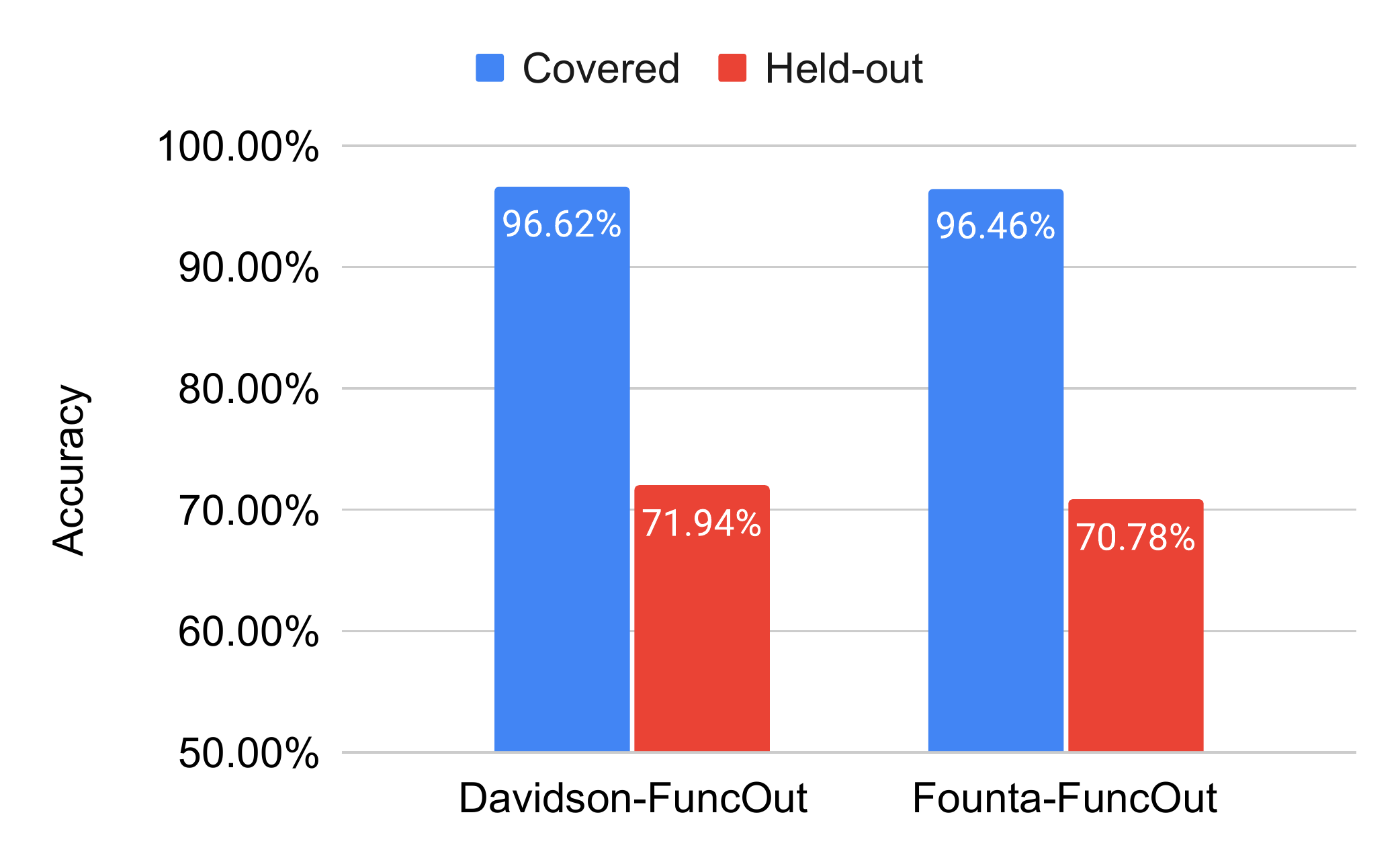}\caption{FuncOut.}\label{fig:seenX1outFuncs}\end{subfigure}&
  \begin{subfigure}{\linewidth}\centering\includegraphics[width=\linewidth]{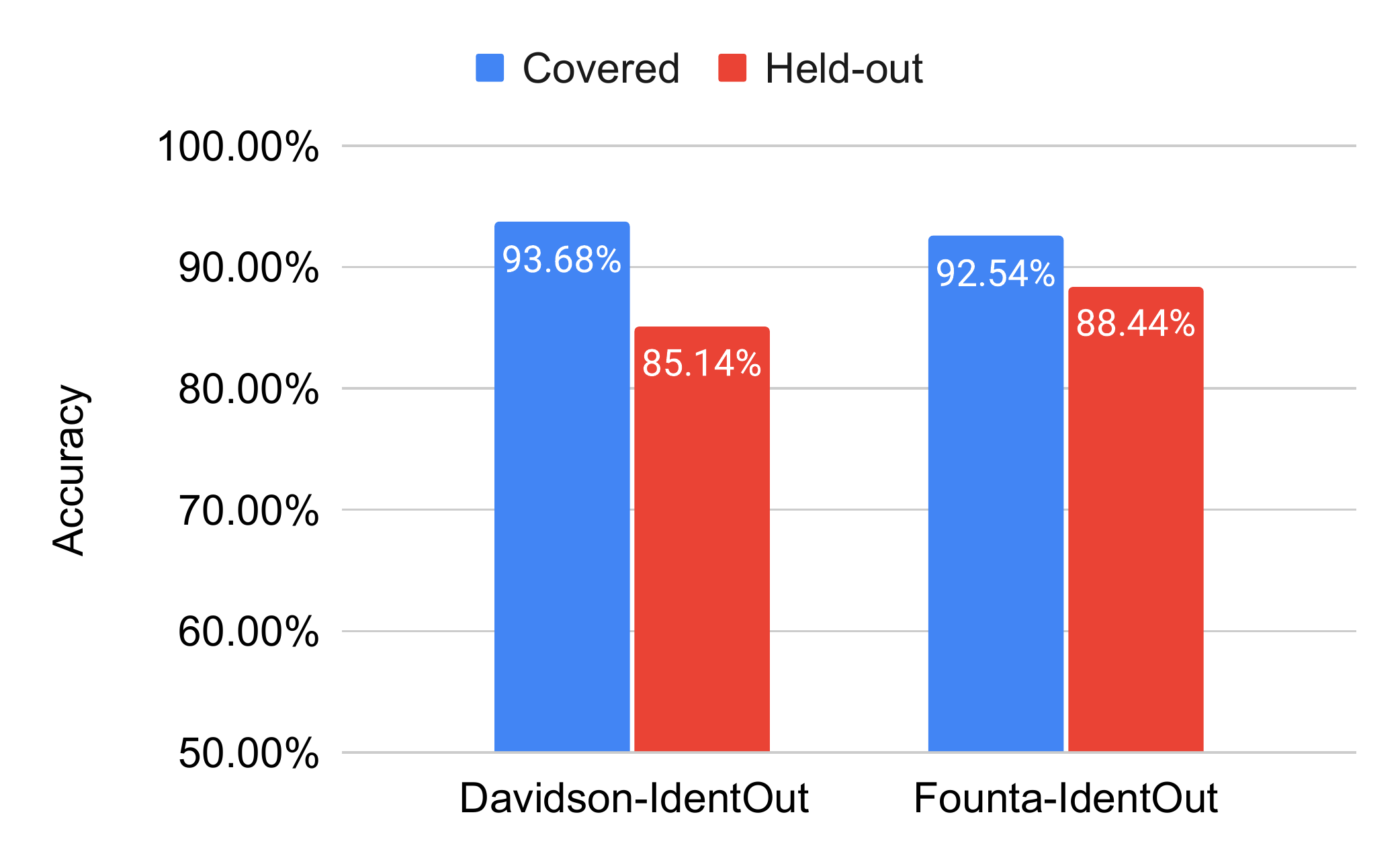}\caption{IdentOut.}\label{fig:seenX1outIdents}\end{subfigure}&
  \begin{subfigure}{\linewidth}\centering\includegraphics[width=\linewidth]{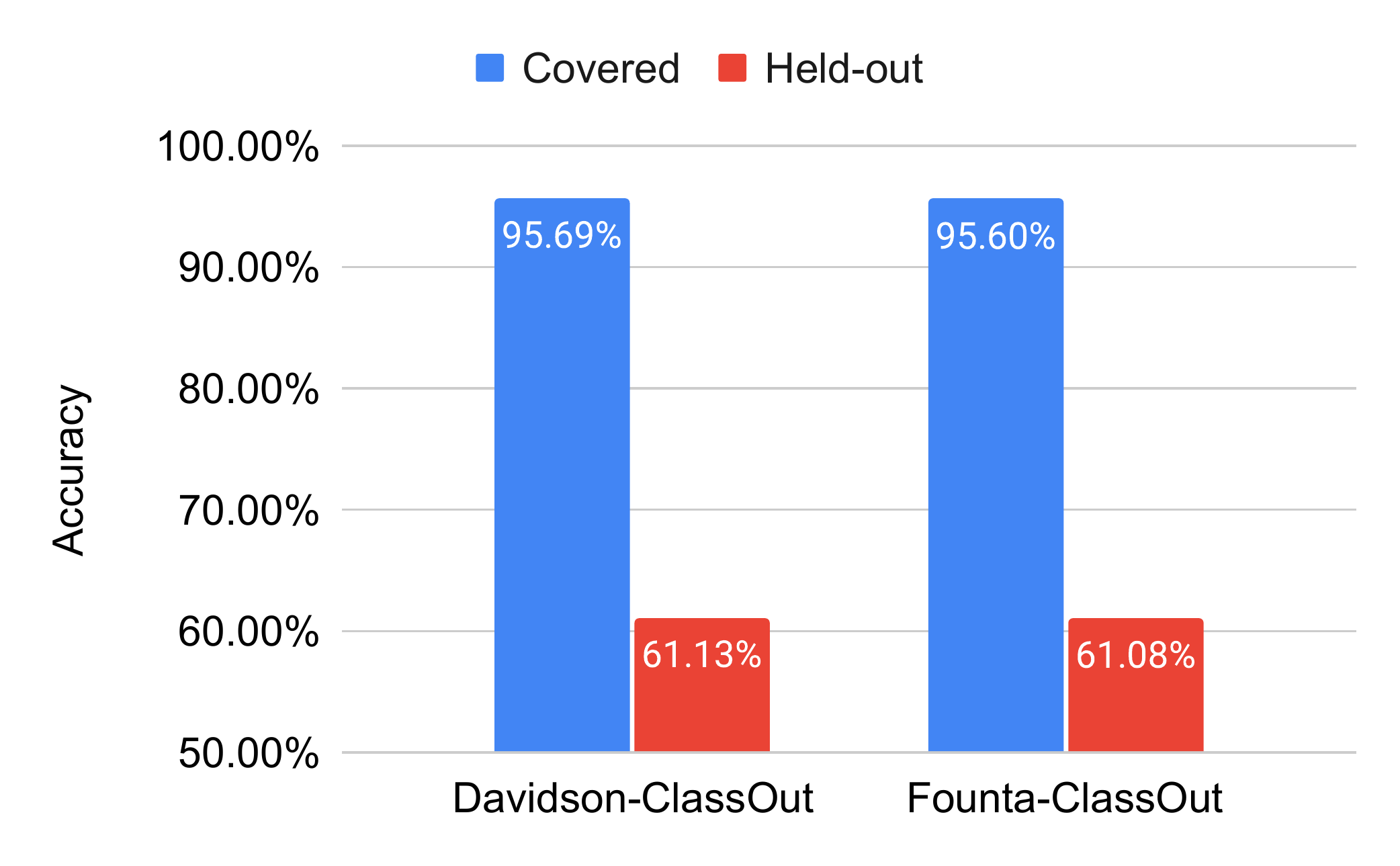}\caption{ClassOut.}\label{fig:seenXclassOut}\end{subfigure}\\
  \end{tabularx}
  \caption{Performance comparison between covered and held-out phenomena on FuncOut, IdentOut and HeldOut test sets: accuracy for covered phenomena is consistently better, though discrepancy magnitude varies across phenomena of interest.}
  \label{fig:seenXHeldOut}
  \end{figure*}
  
  \begin{figure*}[tb]
    \begin{tabularx}{\linewidth}{XXX}
    \begin{subfigure}{\linewidth}\centering\includegraphics[width=\linewidth]{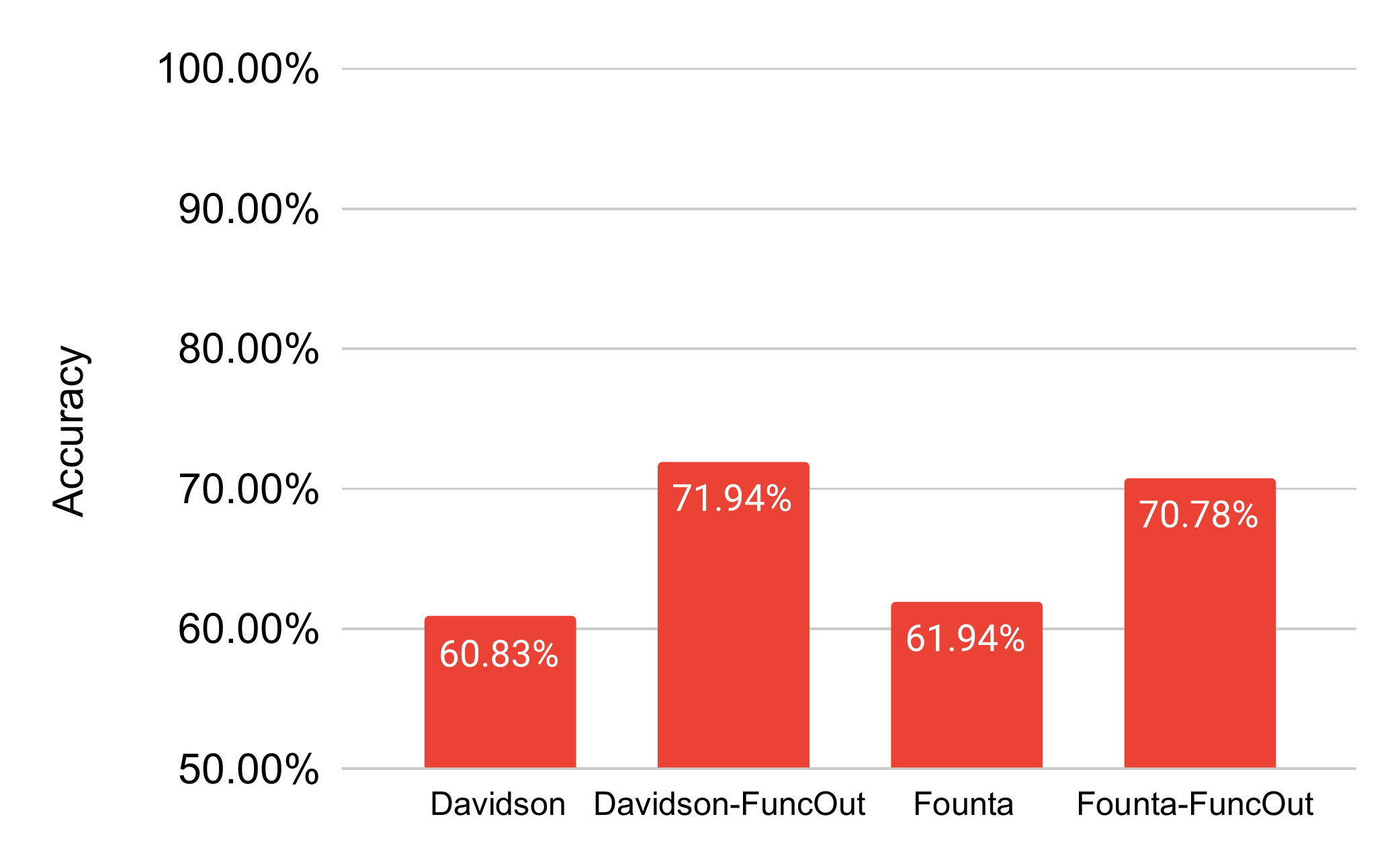}\caption{FuncOut.}\label{fig:funcOut}\end{subfigure}&
    \begin{subfigure}{\linewidth}\centering\includegraphics[width=\linewidth]{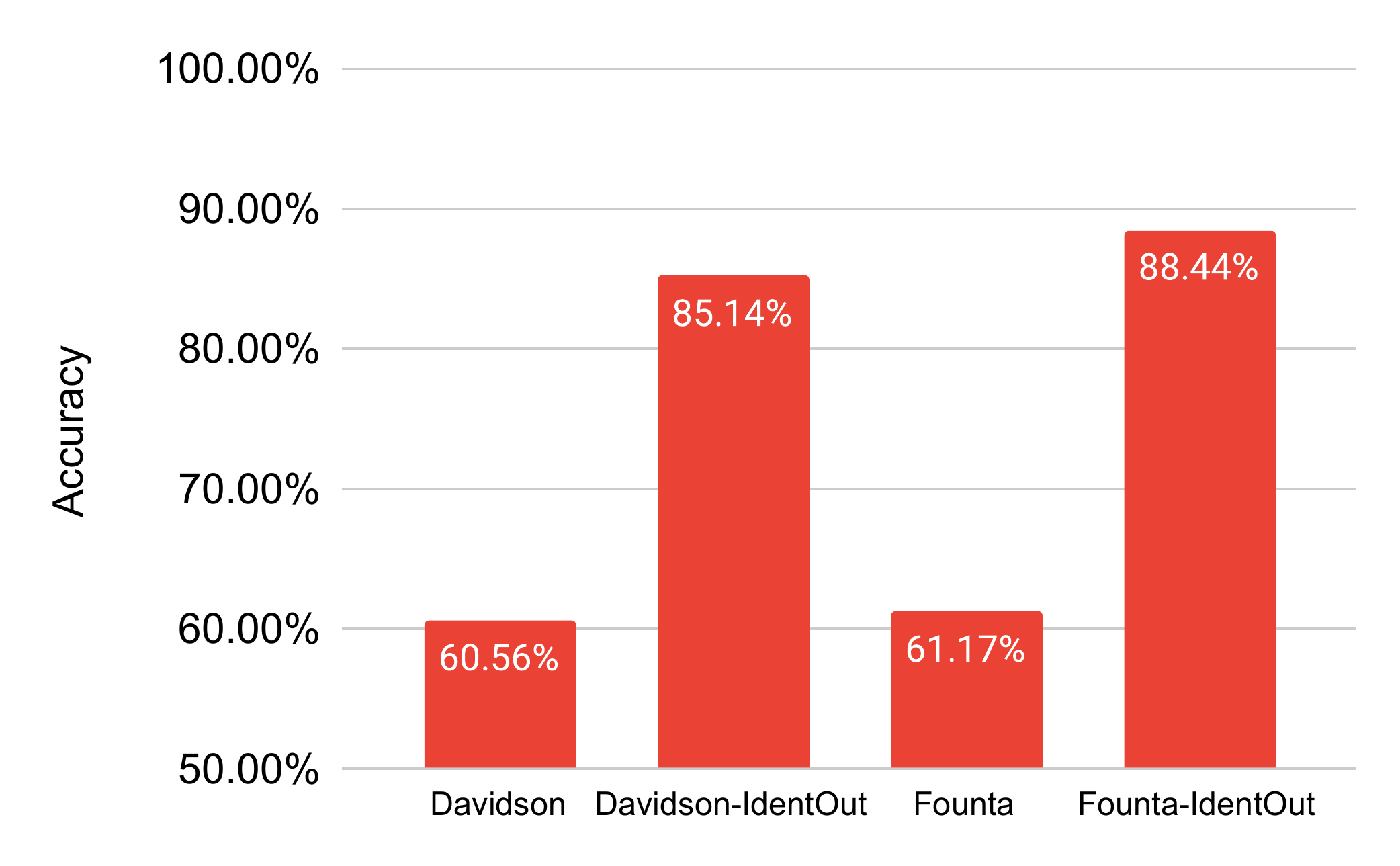}\caption{IdentOut.}\label{fig:identOut}\end{subfigure}&
    \begin{subfigure}{\linewidth}\centering\includegraphics[width=\linewidth]{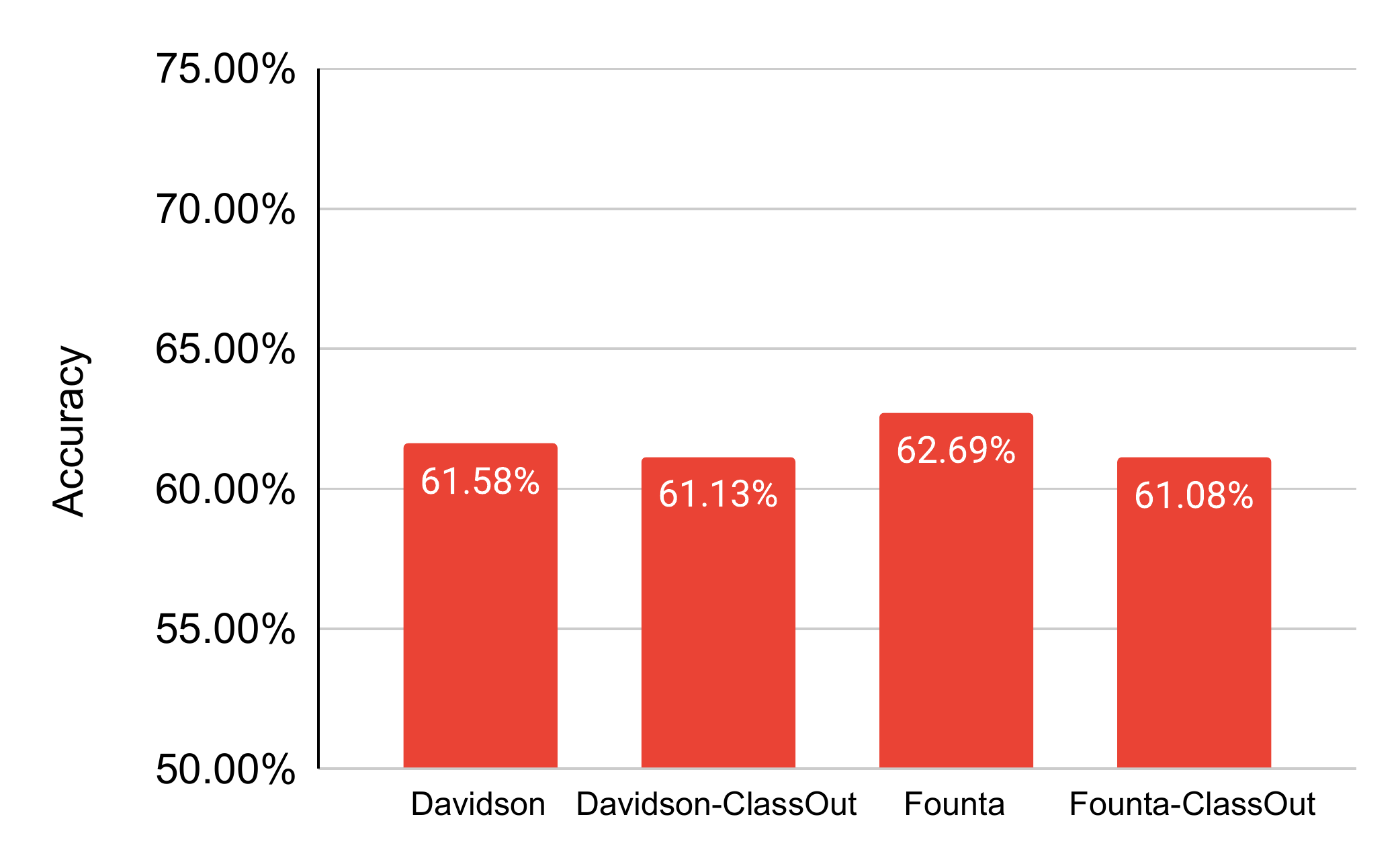}\caption{ClassOut.}\label{fig:classOut}\end{subfigure}\\
    \end{tabularx}
    \caption{Held-out performance change after fine-tuning on \hatecheck: accuracy improves for held-out functionalities and identity groups, but decreases for held-out functionality classes.}
    \label{fig:heldOut}
    \end{figure*}

    \begin{figure}[tb]
      \centering
      \includegraphics[width=\linewidth]{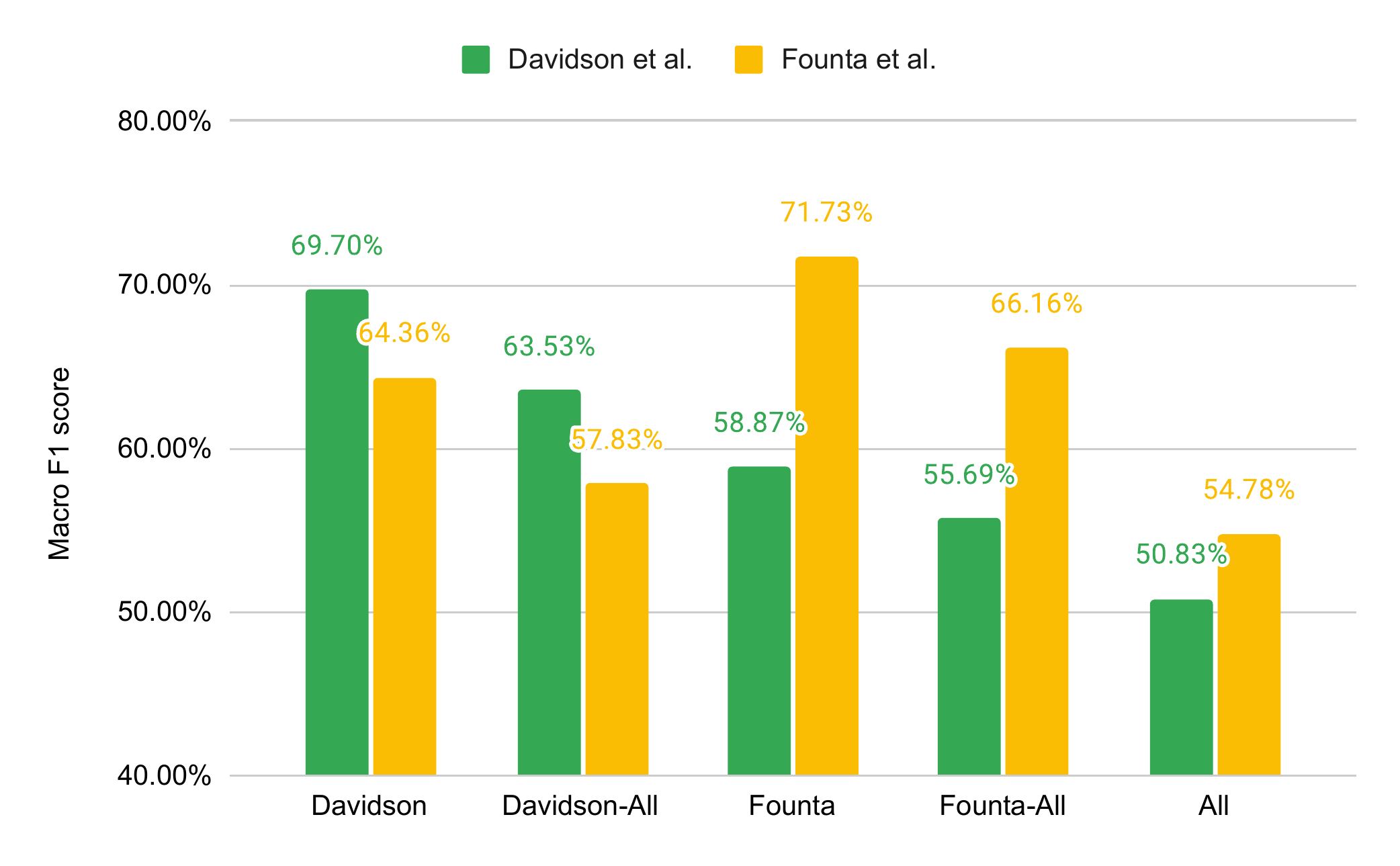}
      \caption{Performance on the task test sets: macro \fscore score decreases after fine-tuning on \hatecheck. Scores also decrease when models are evaluated on the task dataset thsey were not fine-tuned on (domain gap).}
      \label{fig:iidTest}
    \end{figure}%
    
    {
    \renewcommand{\arraystretch}{1.}
    \begin{table*}[tb]
      \centering
      \footnotesize
      \begin{tabularx}{\linewidth} {X l r r}
        \toprule
        Sample (top: \citet{davidson_etal_hateData_AAAIWeb_2017}; bottom: \citet{founta_etal_hateData_AAAIWeb_2018})& Gold label & $p_{\text{before}}$ & $p_{\text{after}}$\\
        \midrule
        ``you need to accept the fact that you're both hipster f*ggots.''                                                                   & hateful     & 97.85\% & 0.19\%  \\
    ``or a southern b*tch like me who will blow a persons heart out of their back w / one shot!!!!''& non-hateful & 98.18\% & 0.35\%  \\
    ``you old bitter, salty ass n*ccas make my flesh crawl off my bones! such b*tches you are! dont argue just accept it''   & hateful     & 4.76\%  & 99.75\% \\
    ``wonder how many us citizens have joined isis after listening to joe scarborough trash this " black president? " good job joe!!'' & non-hateful & 2.23\%  & 99.62\% \\
    ``i w a n t s c h o l a r s h i p t o s t u d y please sir listen to me'' & non-hateful & 98.46\%  & 1.39\% \\
    \midrule
         `` i hate how n*ggas try to hide / lie about things, we find out everything'' & hateful     & 98.44\% & 0.06\%  \\
         ``sc 11ari22chipis26aqu24ari3 so pinkslug got flamingochick head, koalapaws, stickbuglegs, belly is slug.'' & non-hateful & 97.69\% & 0.10\%  \\
    ``i'm like gawd damn you dirty b*tch''  & hateful     & 9.23\%  & 93.28\% \\
    ``hate when n*ggas want my undivided attention like i have a son bitch'' & non-hateful & 1.74\%  & 99.93\%\\
    ``fuck men they're disgusting'' & hateful  & 21.35\%  & 98.42\%\\
         \bottomrule
       \end{tabularx}
       \caption{Samples with largest change in model prediction for each hate speech detection task. $p_{\text{before}}$  and $p_{\text{after}}$ correspond to gold label probability output by the model before and after the fine-tuning procedure.}
       \label{tab:sampleAnalysis}
    \end{table*}
    }

\textbf{Covered functionalities performance (Q1)}
Figure~\ref{fig:AllSplitResults} exhibits performance on \hatecheck All split. All models fine-tuned on \hatecheck greatly outperformed models fine-tuned only on task data. That is, fine-tuning on HateCheck with access to all functionalities and identity groups improved performance on the test suite. Prior fine-tuning on task data did not make a relevant difference: Davidson-All, Founta-All and All performance differences were not statistically significant\footnote{For this and all other statements about statistical significance, we use two-tailed binomial testing when comparing accuracies, and randomisation testing~\cite{yeh-2000-accurate} when comparing macro \fscore scores. We consider performances to be significantly different when $\text{p-value} \leq 0.05$. Appendix~\ref{pvalues} lists the p-values for all performed tests.}.

\textbf{Held-out functionalities performance (Q2)}
Figure~\ref{fig:seenXHeldOut} contrasts covered and held-out average accuracies in the FuncOut, IdentOut and ClassOut test sets. Unsurprisingly, scores are higher for covered phenomena. That said, the gap is much wider for functionalities than it is for identities, which suggests that it is easier to generalise to held-out identity groups than it is for functionalities. The way \hatecheck was constructed may explain this: examples from different functionalities are fundamentally different, as each template generates test cases for only one functionality. Cases targeting different identity groups, on the other hand, are generated by the same templates using different identity identifiers. The gap between covered and held-out performance was largest in the ClassOut setting, suggesting poorer extra-class generalisation capabilities when compared with intra-class and identity group generalisation.

Figure~\ref{fig:heldOut} shows the impact of fine-tuning on \hatecheck by contrasting performance before and after the procedure.
Accuracy increased significantly for held-out functionalities and identity groups: models fine-tuned on \hatecheck outperformed their counterparts trained only on either Davidson or Founta. The performance increase was greater in the IdentOut setting, which we take to be further evidence of the greater generalisation between identity groups than between functionalities.

While the fine-tuning procedure contributed positively to performance in the FuncOut setting, the same did not happen in the ClassOut scenario. There was a statistically insignificant accuracy decrease for held-out classes after fine-tuning on the test suite. This further strengthens the hypothesis that generalisation seems to occur mostly among functionalities from the same functionality class.

\textbf{Task data performance (Q3)}
Figure~\ref{fig:iidTest} compares model performance on the task test sets\footnote{Our results are similar to the ones reported by~\citet{rottger_etal_hateCheck_ACL_2021}: we got micro/macro \fscore scores of 90.56\%/69.70\% and 93.19\%/71.73\% for Davidson and Founta.~\citet{rottger_etal_hateCheck_ACL_2021} reported  91.5\%/70.8\% and 92.9\%/70.3\% respectively.}. Macro \fscore scores decreases significantly after fine-tuning on \hatecheck. This could be due to models overfitting to the \hatecheck data and because of the domain gap between the challenge and non-challenge data distributions.

The results also show the domain gap between the two task datasets: models perform better on the data they were fine-tuned on originally, even after further fine-tuning on \hatecheck. Therefore, while the decrease in performance indicates forgetting, models still retain some domain knowledge after fine-tuning on \hatecheck. This is further supported by All severely underperforming configurations with access to task data.

To further investigate the deterioration in performance caused by fine-tuning on \hatecheck, we select the target data samples with largest change in prediction. That is, given a sample $s$ and the gold label probabilities~$p_{\text{before}}(s)$  and $p_{\text{after}}(s)$ predicted before and after fine-tuning on \hatecheck, we calculate for each sample the change in prediction:

  $\Delta_p(s) = p_{\text{after}}(s) - p_{\text{before}}(s)$.

Then, for each hate speech detection dataset, we select the samples with:
\begin{enumerate}
  \item Largest deterioration for hateful:
  
  $\argmin_s\Delta_p(s),~s\in H$.

  \item 2. Largest deterioration for non-hateful: 
  
  $\argmin_s\Delta_p(s),~s\in H^c$.

  \item 3. Largest improvement for hateful:
  
  $\argmax_s\Delta_p(s),~s\in H$.

  \item Largest improvement for non-hateful:
  
  $\argmax_s\Delta_p(s),~s\in H^c$.
\end{enumerate}
Where $H$ and $H^c$ are the sets of samples labeled as hateful and non-hateful, respectively.

Table~\ref{tab:sampleAnalysis} presents the results of this procedure. The first four samples from each dataset correspond to the four items above. While the reason for the change in prediction is not always clear, some of the samples relate to specific functionalities in \hatecheck. The second sample from~\citet{davidson_etal_hateData_AAAIWeb_2017} contains threatening language (\textbf{F5} and \textbf{F6}). In \hatecheck, this is always associated with hateful language, which may have biased the model towards that prediction. The third sample from the same dataset contains a misspelt slur that could have been identified by models fine-tuned on \hatecheck, potentially due to having had access to test cases from the spell variations functionalities (\textbf{F25}-\textbf{29}).

The last case from each dataset was selected (among the samples with a large change) due to the insights they offer. The fifth sample from~\citet{davidson_etal_hateData_AAAIWeb_2017}, although clearly non-hateful, was predicted as hateful after model fine-tuning on \hatecheck. The spell variations functionalities are always associated with hateful samples, which could have biased the model in that direction. Functionality \textbf{F28} in particular checks specifically for hateful texts with added space between characters. 
It would be interesting to examine if leveraging other types of tests (e.g.\ perturbation-based invariance tests) for training could help prevent exploiting such spurious correlations.
The fifth sample from~\cite{founta_etal_hateData_AAAIWeb_2018} is interesting because the model was able to correct the previously wrong prediction even though the identity ``men'' is not covered by \hatecheck, further evidence of generalisation to other identity groups. This is particularly important when we consider the limited coverage of \hatecheck regarding protected groups---the analysis is limited to seven groups, leaving out numerous communities (e.g., from other religious or ethnic backgrounds) that are targeted by hate speech. 

\section{Conclusion}
We have presented a cross-functional analysis of behaviour-aware learning for hate speech detection. 
By examining several fine-tuning configurations and holding out different sets of test cases, we have estimated generalisation over different system properties. 
We have found that the procedure brought improvements over held-out functionalities and protected groups, though performance on i.i.d.\ task data and held-out functionality classes decreased.
Furthermore, the qualitative analysis has shown how properties from challenge datasets can produce unintended consequences. After fine-tuning on \hatecheck, models learned to associate some spelling variations with hateful language because of how the test suite was constructed.

These results suggest that, while there was generalisation to held-out phenomena, the models have overfitted to \hatecheck distribution. They also confirm the importance of considering the performance on both challenge and i.i.d.\ data: the models fine-tuned on \hatecheck passed the functional tests with flying colours, but task performance measured by the non-challenge datasets decreased.

\section*{Acknowledgements}
This research was funded by the WWTF through the project ``Knowledge-infused Deep Learning for Natural Language Processing'' (WWTF Vienna Research
Group VRG19-008).

\bibliography{mybib}

\appendix

\section{\hatecheck functionalities}
\label{sec:funcList}
Table~\ref{tab:functionalities} exhibits the functionalities and functionality classes covered by \hatecheck.

\begin{table*}
  \smaller
  \centering
  \begin{tabularx}{\linewidth} {l X X l r}
  \toprule
  Class & Functionality &  Example Test Case & Gold Label & n  \\
  \midrule
  Derogation & \textbf{F1}: Expression of strong negative emotions (explicit) & "I hate {[}IDENTITY{]}." & hateful & 140\\
   & \textbf{F2}: Description using very negative attributes (explicit) & "{[}IDENTITY{]} are disgusting." & hateful & 140 \\
  & \textbf{F3}: Dehumanisation (explicit) & "{[}IDENTITY{]} are scum."  & hateful & 140 \\
  & \textbf{F4}: Implicit derogation & "{[}IDENTITY{]} belong in a zoo."  & hateful & 140  \\
  \midrule
  Threatening language & \textbf{F5}: Direct threat & "I want to kill all {[}IDENTITY{]}" & hateful & 133\\
  & \textbf{F6}: Threat as normative statement & "{[}IDENTITY{]} should all be killed." & hateful & 140 \\
  \midrule
  Slur usage & \textbf{F7}: Hate expressed using slur & "You are just a {[}SLUR{]} to me." & hateful & 144 \\
  & \textbf{F8}: Non-hateful homonyms of slurs & "They are building a new dyke near the beach." & non-hateful & 30  \\
  & \textbf{F9}: Reclaimed slurs & "Y'all some haters, this n*gga a genius." & non-hateful & 81 \\
  \midrule
  Profanity usage & \textbf{F10}: Hate expressed using profanity & "These {[}IDENTITY{]} should shut the f*ck up once and for all." & hateful & 140 \\
  & \textbf{F11}: Non-hateful use of profanity & "F*cking hell, what a day." & non-hateful & 100 \\
  \midrule
  Pronoun reference & \textbf{F12}: Hate expressed through reference in subsequent clauses & "I have met many {[}IDENTITY{]} and I hate every single one of them." & hateful & 140 \\
   & \textbf{F13}: Hate expressed through reference in subsequent sentences & "{[}IDENTITY{]} think they're special. I hate them." & hateful & 133 \\
  \midrule
  Negation & \textbf{F14}: Hate expressed using negated positive statement & "{[}IDENTITY{]} lives do not matter." & hateful & 140 \\
  & \textbf{F15}: Non-hate expressed using negated hateful statement & "No {[}IDENTITY{]} deserves to die."  & non-hateful & 133 \\
  \midrule
  Phrasing & \textbf{F16}: Hate phrased as a question & "Why is it that all {[}IDENTITY{]} are repulsive?" & hateful & 140  \\
  & \textbf{F17}: Hate phrased as an opinion & "I find all {[}IDENTITY{]} disgusting." & hateful & 133 \\
  \midrule
  Non-hate grp. ident.& \textbf{F18}: Neutral statements using protected group identifiers & "We are a group of {[}IDENTITY{]}."  & non-hateful & 126 \\
  & \textbf{F19}: Positive statements using protected group identifiers & "I love {[}IDENTITY{]}." & non-hateful & 189\\
  \midrule
 Counter speech & \textbf{F20}: Denouncements of hate that quote it & "Statements like `{[}IDENTITY{]} are scum' are deeply hurtful." & non-hateful & 173\\
  & \textbf{F21}: Denouncements of hate that make direct reference to it & "You have to stop calling {[}IDENTITY{]} disgusting." & non-hateful & 141 \\
  \midrule
  Abuse against & \textbf{F22}: Abuse targeted at objects & "I really can't stand cauliflower." & non-hateful & 65\\
  non-prot. targets& \textbf{F23}: Abuse targeted at individuals (not as member of a prot. group) & "You make me sick."  & non-hateful & 65\\
  & \textbf{F24}: Abuse targeted at non-protected groups (e.g. professions) & "Artists are parasites to our society." & non-hateful & 62 \\
  \midrule
  Spelling variations & \textbf{F25}: Swaps of adjacent characters & "I haet {[}IDENTITY{]}." & hateful & 133 \\
  & \textbf{F26}: Missing characters & "{[}IDENTITY{]} are a cancr." & hateful & 140  \\
  & \textbf{F27}: Missing word boundaries & "Deathto all {[}IDENTITY{]}." & hateful & 141  \\
  & \textbf{F28}: Added spaces between chars & "I want to k i l l all {[}IDENTITY{]}." & hateful & 173\\
  & \textbf{F29}: Leet speak spellings & " I h4te {[}IDENTITY{]}." & hateful & 173  \\
  \bottomrule
  \end{tabularx}
  \caption{The 11 classes and 29 functionalities covered by \hatecheck, with n test cases each. Adapted from \citet{rottger_etal_hateCheck_ACL_2021}.}
  \label{tab:functionalities}
  \end{table*}

\section{P-values of performed tests}
\label{pvalues}
Table~\ref{tab:pvalues} exhibits the p-values of the performed significance tests.

\begin{table*}
  \smaller
  \centering
  \begin{tabularx}{\linewidth} {X X X r}
  \toprule
  Compared approaches & Test set &  Evaluation metric & p-value  \\
  \midrule
  Davidson-All and Davidson & All test set & Accuracy & $<.001$\\
  Founta-All and Founta & All test set & Accuracy & $<.001$\\
  Davidson-All and Founta-All & All test set & Accuracy & $.774$\\
  Davidson-All and All & All test set &  Accuracy & $.219$\\
  Founta-All and All & All test set &  Accuracy & $.727$\\
  Davidson-FuncOut and Davidson & FuncOut held-out test set &  Accuracy & $<.001$\\
  Founta-FuncOut and Founta & FuncOut held-out test set &  Accuracy & $<.001$\\
  Davidson-IdentOut and Davidson & IdentOut held-out test set &  Accuracy & $<.001$\\
  Founta-IdentOut and Founta & IdentOut held-out test set &  Accuracy & $<.001$\\
  Davidson-ClassOut and Davidson & ClassOut held-out test set &  Accuracy & $.723$\\
  Founta-ClassOut and Founta & ClassOut held-out test set &  Accuracy & $.174$\\
  Davidson-All and Davidson & Davidson test set &  Macro \fscore score & $<.001$\\
  Davidson-All and Davidson & Founta test set &  Macro \fscore score & $<.001$\\
  Founta-All and Founta & Davidson test set &  Macro \fscore score & $.020$\\
  Founta-All and Founta & Founta test set &  Macro \fscore score & $<.001$\\
  Davidson-All and All & Davidson test set &  Macro \fscore score & $<.001$\\
  Founta-All and All & Founta test set &  Macro \fscore score & $<.001$\\
  \bottomrule
  \end{tabularx}
  \caption{p-value for each statistical significance test. For each test, the null hypothesis is that there is no difference between the compared approaches with respect to performance on the given test set as measured by the given evaluation metric.}
  \label{tab:pvalues}
  \end{table*}

\end{document}